\documentclass[10pt,twocolumn,letterpaper]{article}

\usepackage[pagenumbers]{wacv} %

\definecolor{wacvblue}{rgb}{0.21,0.49,0.74}
\usepackage[pagebackref,breaklinks,colorlinks,allcolors=wacvblue]{hyperref}
\usepackage{epigraph}
\usepackage{graphicx}
\usepackage{caption}
\usepackage{subcaption}
\usepackage{xcolor}
\usepackage{array}
\usepackage{multirow}
\usepackage{float}
\usepackage{caption}

\usepackage{pifont}

\definecolor{darkgreen}{rgb}{0.0, 0.5, 0.0}
\definecolor{darkred}{rgb}{0.6, 0.0, 0.0}

\usepackage[dvipsnames]{xcolor}
\newcommand{\gcmark}{\textcolor{ForestGreen}{\ding{51}}}%
\newcommand{\rxmark}{\textcolor{BrickRed}{\ding{55}}}%

\def\wacvPaperID{969} %
\def\confName{WACV}
\def\confYear{2026}

\title{$\mathsf{Cin\acute{e}aste}$: A Fine-grained Contextual Movie Question Answering Benchmark}

\author{%
  \parbox{\linewidth}{\centering
    Nisarg A. Shah$^{1,2}$\footnotemark[1] \quad
    Amir Ziai$^{1}$\quad 
    Chaitanya Ekanadham$^{1}$\quad
    Vishal M. Patel$^{2}$\\
    $^{1}$Netflix, Inc.\quad $^{2}$Johns Hopkins University\\
    {\tt\small snisarg812@gmail.com}
  }
}

\begin{document}
\twocolumn[{%
\renewcommand\twocolumn[1][]{#1}%
\maketitle
\begin{center}
    \includegraphics[width=\textwidth]{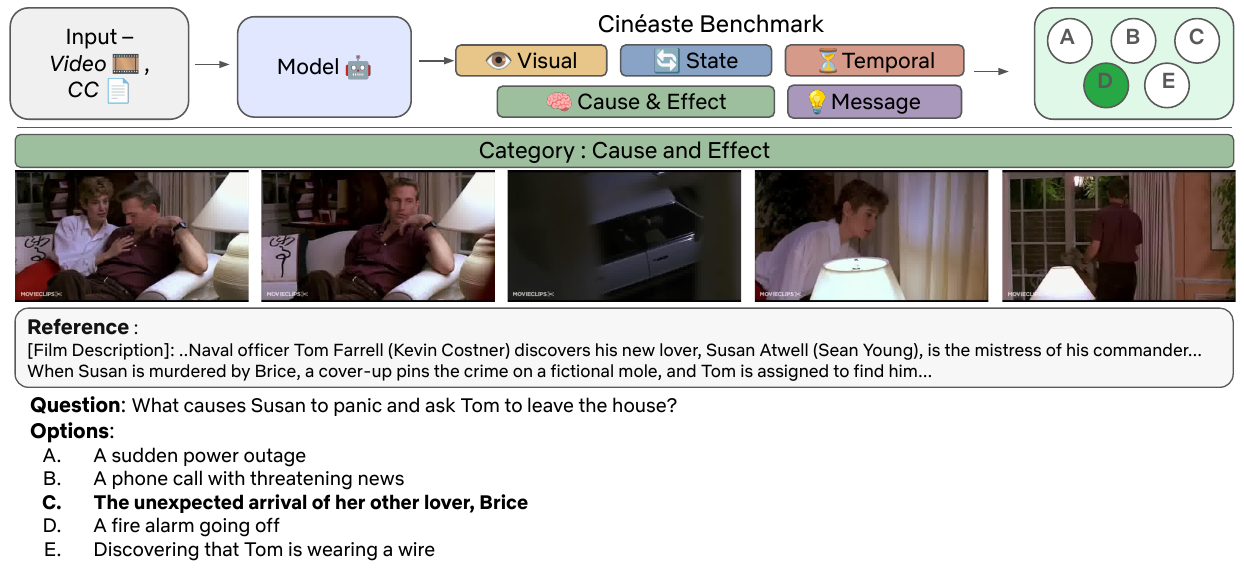} 
    
\captionof{figure}{
    An overview of the \textbf{$\mathsf{Cin\acute{e}aste}$} benchmark and evaluation framework.
    \textbf{(Top)} A multi-modal model is evaluated against the five fine-grained reasoning categories of our benchmark.
    \textbf{(Bottom)} A concrete example from the \textit{Cause and Effect} category, using the film \textit{No Way Out} (1987). All questions undergo a novel two-stage automated filtering process: a Context-Independence filter to ensure visual context is necessary, and a Contextual Veracity filter to remove factual hallucinations.
    \textbf{Insight:} The model must connect multi-scene narrative context (a secret love triangle) with immediate visual cues (a panicked reaction) to understand the character's high-stakes motivation.
}
    \label{fig:teaser}
    \vspace{1em} 
\end{center}
}]
\renewcommand{\thefootnote}{\fnsymbol{footnote}}
\footnotetext[1]{Work done while an intern at Netflix, Inc.}
\renewcommand{\thefootnote}{\arabic{footnote}}

\begin{abstract}
While recent advancements in vision-language models have improved video understanding, diagnosing their capacity for deep, narrative comprehension remains a challenge. Existing benchmarks often test short-clip recognition or use template-based questions, leaving a critical gap in evaluating fine-grained reasoning over long-form narrative content. To address these gaps, we introduce $\mathsf{Cin\acute{e}aste}$, a comprehensive benchmark for long-form movie understanding. Our dataset comprises 3,119 multiple-choice question-answer pairs derived from 1,805 scenes across 200 diverse movies, spanning five novel fine-grained contextual reasoning categories. We use GPT-4o to generate diverse, context-rich questions by integrating visual descriptions, captions, scene titles, and summaries, which require deep narrative understanding. To ensure high-quality evaluation, our pipeline incorporates a two-stage filtering process: Context-Independence filtering ensures questions require video context, while Contextual Veracity filtering validates factual consistency against the movie content, mitigating hallucinations. Experiments show that existing MLLMs struggle on $\mathsf{Cin\acute{e}aste}$; our analysis reveals that long-range temporal reasoning is a primary bottleneck, with the top open-source model achieving only 63.15\% accuracy. This underscores significant challenges in fine-grained contextual understanding and the need for advancements in long-form movie comprehension. 
\end{abstract}

\section{Introduction}

\textit{"Cinema is a matter of what's in the frame and what's out."} \vspace{-2ex} \begin{flushright} - Martin Scorsese \end{flushright} \vspace{-1ex}
Understanding movies requires more than recognizing isolated objects or actions; it involves following complex narratives, character developments, and nuanced interactions that unfold over time. As multi-modal large language models (MLLMs) evolve~\cite{yin2023survey, openai2023gpt4v}, with enhanced capabilities to process both visual and textual information, it becomes essential to assess whether they can interpret stories similarly to human understanding. This necessitates new diagnostic benchmarks designed not just to measure performance, but to reveal specific failure modes in how models grasp the subtle, layered complexity of narrative-driven video content, like movies.

Most existing datasets and models focus on short videos, often under a minute, targeting tasks that require brief temporal understanding~\cite{mangalam2023egoschema, lei2022revealing}. While egocentric videos~\cite{grauman2022ego4d, damen2018scaling, mangalam2023egoschema} and instructional videos~\cite{miech2019howto100m, yang2021just} provide longer sequences, they often lack the narrative depth of movies, focusing instead on immediate actions or procedural tasks. Several movie datasets have been proposed~\cite{rohrbach2016movie, bain2020condensed, tapaswi2016movieqa, lei2019tvqa, soldan2022mad, han2023autoad, wu2021longform, huang2020movienet, vicol2018moviegraphs, xiong2019graphbased, sadhu2021visual, marszalek09, rao2020localtoglobal, chen2022match, bose2022movieclip, maharaj2017dataset, rawal2024cinepile, ghermi2024short}, providing rich narrative content and tasks like clip retrieval, video question answering, audio description, semantic role labeling, and scene segmentation. However, these datasets have limitations, such as short clip durations~\cite{rawal2024cinepile}, reliance on template-based questions~\cite{mangalam2023egoschema,bose2022movieclip,rawal2024cinepile}, and the ability to answer questions using text alone~\cite{lvu2021,ghermi2024short}, which do not fully capture the complexity of long-form movie understanding.

\begin{table*}[t] %
\caption{Comparison of $\mathsf{Cin\acute{e}aste}$ with previous benchmarks. Annotation indicates whether QA pairs are manual, automatic, or mixed. Avg. Length (min) shows the average video duration in minutes. \#QA Pairs lists the total number of question-answer pairs. Multimodal specifies if both video and textual inputs are typically needed for answers. Long-Term denotes if understanding spans multiple scenes (longer than 3 minutes). Accessible reflects the availability of video data. Free-form Questions indicates if questions are open-ended rather than template-bound. Fine-grained Context evaluates detailed scene understanding within the overall narrative. Veracity Filtered indicates if a specific filtering process for factual consistency is used. 'Manual' denotes human-led curation, while our work introduces a scalable, automated approach to this challenge.}
\label{dataset_comparison_table}
\centering
\resizebox{0.85\textwidth}{!}{%
\begin{tabular}{@{}lcrrcccccc@{}}
\toprule
\multirow{2}{*}{\textbf{Dataset}} & \multirow{2}{*}{\textbf{Annotation}} & \textbf{Avg.} & \multirow{2}{*}{\textbf{\#QA Pairs}} & \multirow{2}{*}{\textbf{Multimodal}} & \multirow{2}{*}{\textbf{Long-Term}} & \multirow{2}{*}{\textbf{Accessible}} & \textbf{Free-form} & \textbf{Fine-grained} & \textbf{Veracity} \\ %
& & \textbf{Length (min)} &  & & & & \textbf{Questions} & \textbf{Context} & \textbf{Filtered} \\ %
\midrule
MovieQA~\cite{tapaswi2016movieqa} & Manual & 3.4 & 6,462 & \gcmark & \rxmark & \rxmark & \rxmark & \gcmark & \rxmark \\ %
TGIF-QA~\cite{jang2017tgifqa} & Auto & 0.05 & 25,751 & \rxmark & \rxmark & \gcmark & \gcmark & \rxmark & \rxmark \\ %
MSRVTT-QA~\cite{Xu2017msrvtt} & Auto & 0.25 & 72,820 & \rxmark & \rxmark & \gcmark & \rxmark & \rxmark & \rxmark \\ %
ActivityNet-QA~\cite{yu2019activitynetqa} & Manual & 3 & 8,000 & \rxmark & \rxmark & \gcmark & \rxmark & \rxmark & \rxmark \\ %
TVQA~\cite{lei2019tvqa} & Manual & 1.27 & 15,253 & \gcmark & \rxmark & \rxmark & \rxmark & \rxmark & \rxmark \\ %
How2QA~\cite{sanabria2018how2} & Manual & 1 & 4,400 & \rxmark & \rxmark & \gcmark & \rxmark & \rxmark & \rxmark \\ %
NExT-QA~\cite{xiao2021nextqanext} & Manual & 0.73 & 9,178 & \rxmark & \rxmark & \gcmark & \rxmark & \rxmark & \rxmark \\ %
iVQA~\cite{liu2018ivqa} & Manual & 0.3 & 10,000 & \rxmark & \rxmark & \gcmark & \rxmark & \rxmark & \rxmark \\ %
MoVQA~\cite{zhang2023movqa} & Manual & 16.5 & 4,040 & \gcmark & \gcmark & \rxmark & \gcmark & \rxmark & \rxmark \\ %
EgoSchema~\cite{mangalam2023egoschema} & Manual + Auto & 3 & 5,000 & \rxmark & \rxmark & \gcmark & \rxmark & \rxmark & \rxmark \\ %
MovieChat~\cite{song2023moviechat} & Manual & 7.65 & 2,417 & \gcmark & \gcmark & \rxmark & \rxmark & \gcmark & \rxmark \\ %
CinePile~\cite{rawal2024cinepile} & Manual + Auto & 2.67 & 4,940 & \gcmark & \rxmark & \gcmark & \rxmark & \gcmark & Manual \\ %
SFD~\cite{ghermi2024short} & Manual + Auto & 13.7 & 4,885 & \gcmark & \gcmark & \gcmark & \gcmark & \rxmark & Manual \\ %
Infinibench~\cite{ataallah2024infinibench} & Auto & 53 & 1,600 & \gcmark & \gcmark & \gcmark & \gcmark & \rxmark & \rxmark \\
\midrule
$\mathsf{Cin\acute{e}aste}$ (Ours) & Auto & \textbf{19} & \textbf{3,119} & \gcmark & \gcmark & \gcmark & \gcmark & \gcmark & \gcmark \\ %
\bottomrule
\end{tabular}%
}
\end{table*}

In this work, we introduce $\mathsf{Cin\acute{e}aste}$, a comprehensive benchmark specifically designed for fine-grained contextual movie question answering. Our dataset addresses the aforementioned limitations by focusing on longer sequences of important scenes from movies, averaging around 20 minutes per film—longer than typical short clips (average duration of 2--3 minutes) but shorter than full-length movies (average duration of 120 minutes). This balance provides sufficient narrative context for understanding complex plots while avoiding issues related to full movie distribution and copyright constraints.

A key aspect of $\mathsf{Cin\acute{e}aste}$, is its emphasis on fine-grained contextual understanding. Our benchmark requires models to integrate both visual and textual information across multiple levels of comprehension, challenging them to perform deep narrative reasoning. To achieve this, we designed five fine-grained contextual reasoning categories that encompass various dimensions of movie understanding, progressing from detailed visual analysis to high-level abstract reasoning.
Firstly, \textbf{Visual Reasoning} assesses detailed understanding of visual elements within scenes beyond what is provided in captions. Secondly, \textbf{State Changes} involve tracking transformations in objects or settings over time, requiring temporal reasoning. Thirdly, \textbf{Temporal Ordering} focuses on understanding the sequence of events in the narrative, testing models' ability to comprehend story progression. Moving to deeper analysis, \textbf{Cause and Effect} examines causal relationships between events, demanding reasoning about character motivations and plot dynamics. Finally, at the most abstract level, \textbf{Message Understanding} entails interpreting underlying themes and messages, requiring high-level abstraction and synthesis of narrative elements. These categories are designed to deconstruct the complex task of 'movie understanding' into measurable skills, allowing us to pinpoint exactly where models succeed and fail, a crucial step for guiding future research.

To generate challenging and contextually dependent question-answer pairs aligned with these categories, our QA generation pipeline incorporates comprehensive input modalities—including visual descriptions, closed captions, scene titles, and movie summaries—employing GPT-4o~\cite{chatgpt} for generating nuanced, context-rich questions. To rigorously validate the contextual dependency and factual grounding of each generated QA pair, we employ a two-tiered filtering pipeline. First, a Context-Independent QA Filtering module identifies questions answerable without explicit visual or narrative content. Second, our novel \textit{Contextual Veracity Filter} systematically suppresses hallucinations. The necessity of this step is underscored by our finding that it filters over 25%

Our benchmark comprises 200 movies from diverse genres, totaling 1,805 scenes and approximately 3,119 multiple-choice question-answer pairs across the five reasoning categories. We conduct extensive experiments to evaluate state-of-the-art multi-modal models on $\mathsf{Cin\acute{e}aste}$, revealing significant challenges in fine-grained contextual movie understanding.

In summary, our contributions are:

\begin{itemize}
    \item We introduce $\mathsf{Cin\acute{e}aste}$: a novel benchmark for fine-grained contextual understanding of long-form (~20 min) movie segments. It features QA pairs across five reasoning categories probing deep narrative comprehension, from visual details to themes.

    \item We propose a robust methodology for automated VQA generation using GPT-4o with rich multimodal context. Critically, it incorporates a two-stage filtering pipeline (LLaMa-3.1 based Context-Independence and a novel Contextual Veracity filter) to ensure context-dependency and factual grounding by mitigating hallucinations.

    \item We provide extensive evaluations of state-of-the-art MLLMs on $\mathsf{Cin\acute{e}aste}$. Our analysis quantifies current model limitations, reveals significant challenges particularly in complex reasoning, and offers clear directions for future research in long-form video understanding.

\end{itemize}

By providing this benchmark, we aim to advance the development of models capable of deep narrative understanding over extended video content.

\label{sec:intro}

\section{Related Works}
\subsection{Video Question Answering Benchmarks}
Video Question Answering (VideoQA) benchmarks are crucial for assessing models' ability to reason over video content. Existing datasets often focus on short videos, typically under a minute, targeting tasks that require minimal temporal understanding~\cite{Xu2017msrvtt,jang2017tgifqa,xiao2021nextqanext,grundemclaughlin2021agqa}. Datasets like MSRVTT-QA~\cite{Xu2017msrvtt} and MSVD-QA~\cite{Xu2017msrvtt} provide large collections of automatically generated question-answer pairs but focus on short, descriptive queries without enabling a global understanding of the video. Similarly, ActivityNet-QA~\cite{yu2019activitynetqa}, TVQA~\cite{lei2019tvqa}, and HowTo100M~\cite{miech2019howto100m} primarily address local, segment-based questions, limiting their scope for holistic understanding. Overall, these datasets require reasoning over only a few frames, which is insufficient for evaluating models' capabilities in understanding complex narratives.

\subsection{Long-Video Understanding}

The development of long-context models has highlighted the need for benchmarks that evaluate reasoning over extended video sequences. Recent datasets like EgoSchema~\cite{mangalam2023egoschema} focus on 3-minute egocentric videos with perceptual questions, while others like Video-MME~\cite{fu2024video}, MVBench~\cite{li2024mvbench}, and LVBench~\cite{wang2024lvbench} emphasize temporal reasoning across longer segments. However, the lengths of these videos often make question generation labor-intensive, and they may not capture the narrative complexity of movies.

In the context of movies, Long Video Understanding (LVU)~\cite{wu2021longform} was an early attempt, focusing on tasks like genre classification and view count prediction, often solvable from single frames and thus not requiring deep understanding. MovieQA~\cite{tapaswi2016movieqa} introduced plot-level questions but relied largely on dialogue, limiting the need for visual reasoning. More recent datasets like CinePile~\cite{rawal2024cinepile} and SFD~\cite{ghermi2024short} have made significant strides in movie-based reasoning. CinePile focuses on fine-grained understanding in shorter scenes, while SFD utilizes short films to build a language-focused benchmark. Our work complements these by focusing on longer (~20 min) concatenated scenes from feature films to specifically test long-range temporal and narrative cohesion, a different facet of the overall challenge.

Despite these efforts, there remains a need for benchmarks that use longer, narratively coherent video sequences to probe specific reasoning skills beyond simple recognition, an area our work aims to address. Concurrent work, Infinibench~\cite{ataallah2024infinibench}, pushes the boundary of long-form understanding with videos averaging over 50 minutes. It provides an essential resource for evaluating comprehension at a very large scale. $\mathsf{Cin\acute{e}aste}$ serves as a complementary benchmark, focusing on a different point in the design space: our ~20-minute segments are specifically constructed to form condensed narratives that facilitate targeted, fine-grained diagnostic questions across our five reasoning categories, whereas Infinibench tests broader comprehension over much longer, continuous sequences.

\subsection{Multi-modal Large Language Models}

Advancements in Multi-modal Large Language Models (MLLMs) have improved the integration of visual and textual data~\cite{yin2023survey}. These models typically include a vision encoder, a modality alignment module, and a language model backbone~\cite{touvron2023llama,vicuna2023}. Extending MLLMs to video data poses challenges in modeling temporal sequences~\cite{li2023videochat,damonlpsg2023videollama}. While some models address this by encoding frames and capturing temporal relationships through specialized modules~\cite{damonlpsg2023videollama}, current MLLMs still struggle with long-range dependencies in video content~\cite{li2023mvbench,xu2023mplug2}.

This underscores the need for models designed to efficiently process extended visual data and for benchmarks capable of evaluating long-form video understanding, particularly in the context of complex narratives found in movies. By providing such benchmarks, we can facilitate the development of models that are better equipped to handle fine-grained contextual reasoning over extended video content.

\begin{figure*}
  \centering
  \includegraphics[width=0.8\linewidth]{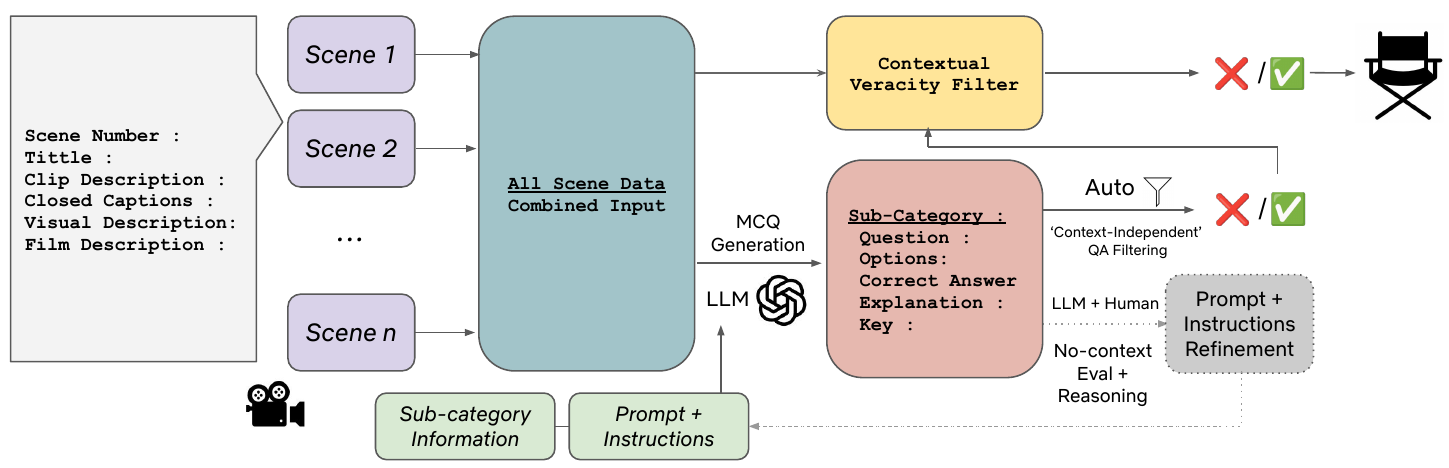}
\caption{Automated QA Generation Pipeline for the $\mathsf{Cin\acute{e}aste}$ Dataset. First, multiple movie-relevant input data — including Scene Number, Title, Clip Description, Closed Captions, Visual Description, and Film Description — are aggregated across scenes to create a comprehensive context. The pipeline operates in two main phases. In Phase 1 (grey-dotted box), QA generation prompts are iteratively refined using an initial 20 movies, incorporating feedback from Large Language Models (LLMs) and human annotators based partly on assessing 'context-independent' answerability. Once prompts demonstrate reliable generation quality, Phase 2 scales the process to all 200 movies, generating questions across five categories. Crucially, following generation in Phase 2, all QA pairs undergo a \textbf{two-stage validation process}: (1) \textit{Context-Independence filtering} removes questions solvable without reference to the video context, and (2) \textit{Contextual Veracity filtering} eliminates pairs that are factually irrelevant with the provided context or based on hallucinations.}   \label{fig:pipeline_cine}
\end{figure*}

\label{sec:formatting}

\section{\textsf{Cinéaste Dataset}}
The construction of the $\mathsf{Cin\acute{e}aste}$ dataset involves three primary stages of dataset development, followed by a comprehensive overview of the dataset’s characteristics: 1) Movie Data Collection and Consolidation, where raw video content and metadata are systematically gathered; 2) Definition of categories for Model Evaluation, which evaluates range of reasoning skills for long-range, movie understanding; 3) Automated QA Generation, which creates question-answer pairs with a refinement process; and finally, 4) Dataset Statistics, providing a quantitative overview into the dataset’s attributes and distributions.

\subsection{Movie Data Collection and Consolidation}
We collected video clips from English-language films available on the YouTube channel MovieClips\footnote{\url{https://www.youtube.com/@MOVIECLIPS}}, which provides scenes capturing significant plot points useful for scene-level question answering\cite{bain2020condensed,rawal2024cinepile}.

To create a dataset suited for long-video (\textgreater 15 min) question answering, focusing on extended reasoning and retrieval, we combined multiple key scenes from the same movie on the channel. On average, approximately 9 scenes were gathered per movie, resulting in about 20 minutes of content per film. This approach using curated, pivotal scenes from a source like MovieClips represents a pragmatic balance, providing rich narrative context essential for complex reasoning tasks while navigating the significant copyright and distribution challenges associated with using full-length films. These scenes are ordered as they appear in the original movies, allowing us to assemble coherent long-form videos akin to condensed versions of the films.

Formally, for each movie \( M \), we have a set of scenes \( \{ S_1, S_2, \dots, S_m \} \). Each scene \( S_x \) contains video frames \( V_x = [v_{x1}, v_{x2}, \dots, v_{x_n}] \) and closed captions \( C_x \). By concatenating these scenes sequentially, we construct the aggregated video \( M' = \bigcup_{x=1}^{m} S_x \).

To enhance scene comprehension, we generated visual descriptions for each scene. For each scene \( S_x \), we uniformly sampled 32 frames from \( V_x \) and employed GPT-4o to create a visual description. These descriptions supplement closed captions by capturing key visual elements, objects, and implicit cues—such as emotions, atmosphere, and scene dynamics—offering a more comprehensive representation of each scene.
Additionally, metadata for each scene was collected, including the scene title and a logline—a brief summary provided by the video creator. For each movie \( M \), we also scraped an overall movie summary \( D_M \), which outlines the plot and serves as input in our QA generation pipeline.

By aggregating scenes, metadata, and visual descriptions, we created a dataset suitable for input in robust movie-level question-answering generation pipleine.

\begin{figure*}[t]
  \centering
  \includegraphics[width=0.8\linewidth]{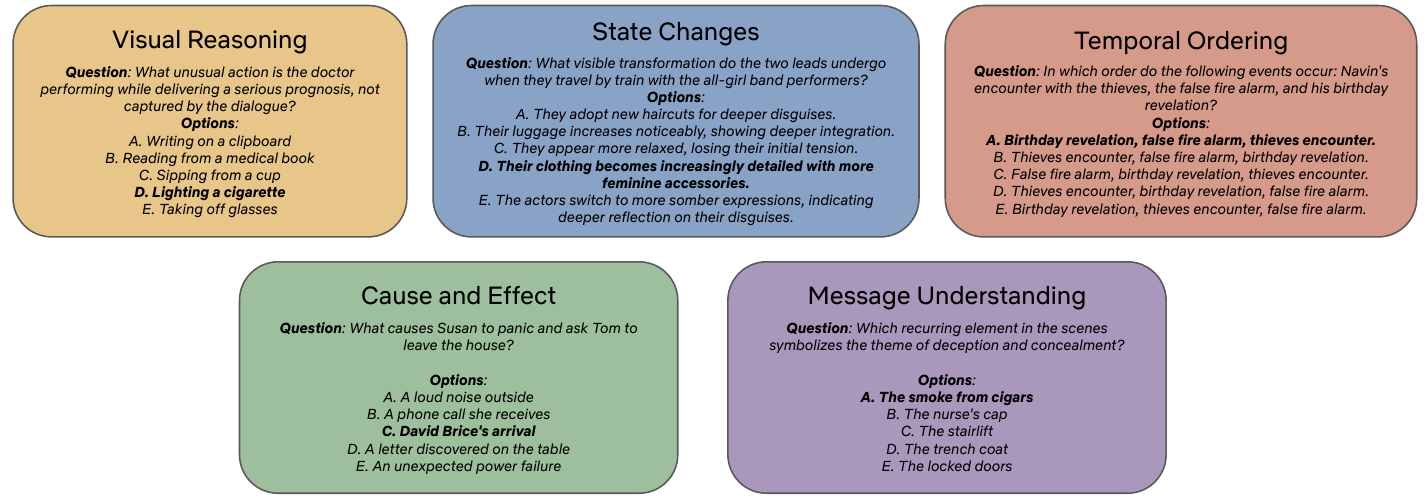}
  \caption{Examples from our five fine-grained reasoning categories. Each question is designed to be unanswerable without specific, multi-modal context from the video, probing skills from detailed visual recognition (Visual Reasoning) to abstract thematic synthesis (Message Understanding). Correct answers are highlighted in green.}
  \label{fig:sample_cine}
\end{figure*}

\subsection{Creation of categories for Model Evaluation}

In current VideoQA benchmarks, question generation often depends on human annotators, allowing creative freedom but lacking scalability due to high time and cost \cite{tapaswi2016movieqa}. Template-based approaches address scalability by guiding annotators with predefined structures \cite{xiao2021nextqanext,lei2018tvqa,patraucean2024perception,rawal2024cinepile}, though they may limit question diversity and depth.

To balance scalability and richness in question diversity, we defined five categories that emphasize essential reasoning and retrieval skills for long-video understanding. These categories were formulated by considering key aspects that professionals in the film industry focus on during analysis, ensuring that questions are varied and meaningful without being constrained by rigid templates. 

Our categories starts with (finding the relevant scene and) detailed analysis at the scene level and progressively expands to encompass broader narrative elements. This structure allows us to evaluate models across a spectrum of understanding, from fine-grained visual recognition to the interpretation of overarching themes.

As shown in Fig.~\ref{fig:sample_cine}, the five categories are:

\textbf{Visual Reasoning}: Requires recognizing and interpreting visual elements within a scene, such as objects or environmental details not mentioned in captions.

\textbf{State Changes}: Involves tracking transformations in objects, settings, or characters over time, both within and across scenes.

\textbf{Temporal Ordering}: Assesses understanding of the plot's progression by focusing on the sequence and relationship of narrative events.

\textbf{Cause and Effect}: Requires analyzing causal relationships between events. Questions involve identifying how one event leads to another or understanding character motivations, demanding deeper reasoning.

\textbf{Message Understanding}: Targets grasping underlying messages, themes, or symbols in the narrative. Questions involve identifying thematic elements and understanding their significance throughout the video.

This structured benchmark enables diverse question generation, ensuring comprehensive model evaluation on long-form content without rigid constraints, achieving a balance between scalability and question depth.

\subsection{QA Generation Pipeline with LLMs}

To create $\mathsf{Cin\acute{e}aste}$  for long-form and movie-level question answering, our QA generation pipeline uses GPT-4o~\cite{chatgpt} to generate questions that assess long-term comprehension and detailed reasoning across movie content. Later, we use automatic no-context filtering to ensure benchmark have relevant and contextual questions

\textbf{Automatic QA Generation.} 
To generate QA, we use a structured format that consolidates all curated information for each movie, including scene-level details such as visual descriptions, closed captions, loglines, scene titles, and the overarching film description. This compilation of inputs across scenes provides a comprehensive context, ensuring that questions require a nuanced understanding of both individual scenes and the overall narrative.

For each movie, we prompt the model separately for each question sub-category, providing specific instructions that match the sub-category’s goals. Each prompt includes a description of the sub-category and guidelines for creating high-quality questions and distractors. This setup helps GPT-4~\cite{chatgpt} generate questions that accurately test the intended skills and avoid common template issues like predictability or reliance on general knowledge.
The output consists of questions in JSON format, each with a correct answer, and four distractors designed to be plausible but slightly misleading, requiring attention to the specific visual and textual data. Scene titles are referenced indirectly to provide context without explicitly revealing scene numbers.

Alongside the correct answer, each output from the LLM includes the reasoning behind the answer and a "Key" identifying the relevant scenes. This additional information can supports training reasoning models by detailing the logical path required to reach accurate answers\cite{GPT-4o,gemini_flash}.

This QA pipeline generated 11,729 questions across five categories, covering 200 movies and 1,805 scenes. By using detailed prompts and integrating scene-level data, this approach overcomes the limitations of template-based generation, resulting in a dataset that effectively evaluates long-form, story-driven movie underestanding.

\textbf{Context-Independent QA Filtering.} 
Once the QA pairs are generated, it’s crucial to filter out questions that might be answerable without referring to the video or captions, a limitation noted in prior studies~\cite{tapaswi2016movieqa,rawal2024cinepile,ghermi2024short}. These “Context-Independent” questions can often be answered through general reasoning and subtle hints within the question and options alone, bypassing the need for video-specific information. To test for this, we evaluate each question by presenting only the question and options to the model—without access to the movie content—and assess if it can predict the correct answer reliably.

Initially, we explored refining generated questions via iterative regeneration through LLMs. However, preliminary experiments indicated that such regeneration frequently compromised the original precision and reasoning complexity, producing overly generalized or ambiguous questions. This iterative method also introduced computational overhead without substantial improvements in eliminating context-independent questions. Consequently, we revised our initial prompting strategy to proactively minimize context independence.

Addressing these constraints, we established a robust filtering criterion leveraging multiple independent predictions to identify questions reliably answerable through general reasoning alone. Specifically, we conducted evaluations with LLaMa-3.1-70B~\cite{dubey2024llama}, prompting the model with five independent seeds. Questions correctly answered in at all five out of five trials (= 100\% accuracy) without video-specific context were identified and subsequently removed. This methodological refinement substantially increased the integrity and contextual necessity of retained questions.

\textbf{Contextual Veracity Filter for Hallucination Mitigation}
Despite carefully constructed prompts and in-context learning examples in the automated QA generation pipeline, we encountered hallucinations in GPT-4o outputs. Hallucinations typically emerged as questions referencing nonexistent visual attributes, incorrect spatial relations, inferred but unsupported narrative elements, or speculative character intentions and interactions. Although explicit prompt adjustments partially mitigated these issues, residual hallucinations persisted, necessitating an additional verification layer.

To systematically detect and eliminate these hallucinations, we introduce the \textit{Contextual Veracity Filter}, a textual reasoning-based validation module leveraging the capabilities of LLaMa-3.1-70B~\cite{dubey2024llama}. For each question-answer pair $(Q, A)$, the module receives comprehensive textual context $C = \{V_{desc}, C_{cap}, S_{title}, D_M, S_{logline}\}$ alongside the question and answer options. Formally, the validation procedure can be expressed as:
\begin{equation}
    \mathcal{V}(Q, A, C) = \frac{1}{N}\sum_{i=1}^{N}\mathbb{I}[\mathsf{LLaMa}(Q, C; \theta_i) = A]
\end{equation}
where $\mathbb{I}[\cdot]$ is the indicator function verifying the correctness of LLaMa's predicted answer, and $\theta_i$ represents different random seeds for the decoding process to ensure robustness against model stochasticity. We set $N=5$ to achieve a reliable estimate of the validation consistency.

We adopt a conservative acceptance criterion, retaining only those question-answer pairs for which $\mathcal{V}(Q, A, C) \geq 0.8$. This ensures that questions are reliably answerable using provided contextual information alone, significantly enhancing the dataset’s reliability and coherence.

Such a rigorous filtering approach is particularly beneficial when generating large-scale benchmarks automatically, as it ensures the contextual integrity and relevance of generated questions, ultimately improving downstream model evaluations.

\begin{figure}[ht]
    \centering
    \resizebox{1.0\linewidth}{!}{%
        \begin{subfigure}{0.45\textwidth}
            \centering
            \includegraphics[width=\linewidth]{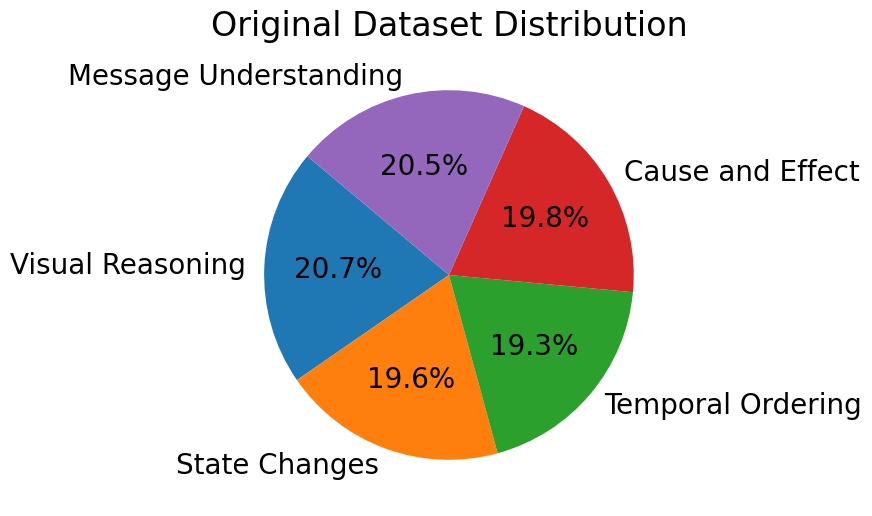}
            \label{fig:original_distribution}
        \end{subfigure}
        \quad
        \begin{subfigure}{0.45\textwidth}
            \centering
            \includegraphics[width=\linewidth]{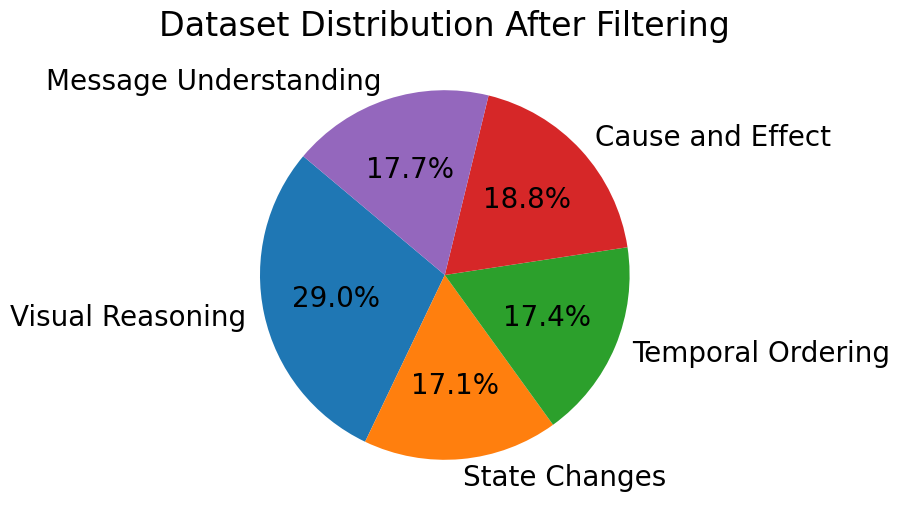}
            \label{fig:filtered_distribution}
        \end{subfigure}
    }
    \caption{Comparison of Dataset Distributions Before and After Filtering}
    \label{fig:dataset_distributions}
\end{figure}

\begin{table*}[htp]
\centering
\caption{Performance of the evaluated models on the $\mathsf{Cin\acute{e}aste}$ benchmark, measured by multiple-choice accuracy. The acronyms for each category are as follows: \textbf{VR} - Visual Reasoning, \textbf{SC} - State Changes, \textbf{TO} - Temporal Ordering, \textbf{CE} - Cause and Effect, and \textbf{MU} - Message Understanding. Models with parameters (-) are closed-source model, whereas all other models in the table are open-source.}
\label{tab:vlm_eval}
\renewcommand{\arraystretch}{1.2}
\scalebox{0.85}{
\begin{tabular}{@{}l|c|c|ccccc|c@{}}
\toprule
\textbf{Model} & \textbf{Parameters} & \textbf{Frames} & \textbf{VR} & \textbf{SC} & \textbf{TO} & \textbf{CE} & \textbf{MU} & \textbf{Avg. Accuracy} \\
\midrule
GPT-4o & -- & 32 & 72.60 & 75.38 & 74.26 & 82.08 & 76.81 & 75.89 \\
Gemini-2.0-Flash & -- & 0.1 \textit{fps} & 74.28 & 66.23 & 54.96 & 66.38 & 65.39 & 66.21 \\
Claude-3.5-Sonnet & -- & 20 & 48.73 & 57.71 & 63.05 & 68.43 & 67.39 & 59.76 \\
\midrule
Aria\cite{li2024aria} & 8x3.5B & 128 & 63.09 & 64.66 & 54.78 & 66.21 & 67.03 & 63.15 \\
LongVA-DPO\cite{zhang2024longva} & 7B & 64 & 51.93 & 49.44 & 39.71 & 55.63 & 55.07 & 50.63 \\
LLaVA-NeXT\cite{li2024llava} & 7B & 64 & 57.02 & 56.02 & 38.05 & 50.68 & 47.10 & 50.59 \\
VideoLLaMA3\cite{zhang2025videollama} & 7B & 64 & 56.35 & 52.26 & 37.68 & 53.24 & 49.28 & 50.56 \\
MiniCPM\cite{hu2024minicpm} & 7B & 64 & 50.17 & 45.68 & 40.07 & 48.29 & 54.35 & 48.03 \\
InternVL2\cite{chen2024expanding} & 7B & 64 & 38.78 & 44.17 & 36.40 & 46.93 & 47.10 & 42.29 \\
ChatUniVi\cite{jin2024chat} & 7B & 64 & 31.18 & 38.21 & 38.56 & 33.69 & 44.04 & 36.41 \\
\bottomrule
\end{tabular}
}
\end{table*}

\subsection{{Cinéaste Dataset Statistics}}
Our benchmark consists of 200 movies, with scenes aggregated to form condensed versions suitable for long-video QA. While the scale of 3,119 QA pairs is primarily intended for robust evaluation, it provides a challenging testbed for existing general-purpose models. On average, each movie includes approximately 9 scenes, with an average scene length of 126 seconds (about 2.1 minutes). This results in an average movie duration of around 19 minutes (1140.7 seconds), totaling approximately 60 hours of content. The rich textual context for benchmark creation includes an average of 240.3 words from captions and 344.5 words from generated visual descriptions per scene. More detailed statistics and distributions of this input data are available in the supplementary material.

The final dataset of 3,119 high-quality QA pairs is the result of a rigorous, multi-stage filtering process. Our automated pipeline initially generated 11,729 candidate questions. From this, the Context-Independence filter first removed 7,544 questions solvable without video context, leaving 4,185 pairs. Subsequently, our Contextual Veracity filter removed an additional 1,066 pairs (25.5\% of the remaining set) flagged as hallucinated. This high rate underscores the unreliability of using LLM generation without such a rigorous, explicit verification step and validates our two-stage filtering pipeline as a crucial methodological contribution.

The final category distribution, shown in Fig.~\ref{fig:dataset_distributions}, reflects the inherent challenges of generating and validating abstract reasoning tasks. {Visual Reasoning} questions, tied to concrete, verifiable details, were more robust to filtering and constitute the largest portion of the dataset at 29\%. Conversely, QA for the other four categories, {State Changes},{Temporal Ordering}, {Cause and Effect}, and {Message Understanding} are even at approximately 17\% each. Their reduced final proportions are a direct result of higher filtering rates, underscoring the significant difficulty of automatically generating complex, abstract questions that remain factually consistent and context-dependent.

\begin{table*}[htp]
\centering
\caption{{Ablation Study of VideoLLaMA3 Model with Increasing Number of Frames.} The table shows the accuracy for various categories of reasoning and understanding, measured by the multiple-choice question performance of the VideoLLaMA3\cite{zhang2025videollama} model across different frame counts (8, 16, 32, and 64). This evaluation is based solely on frame data—without the use of closed captions—to fairly assess the impact of increasing the number of frames.}
\label{tab:videollama3_ablation}
\renewcommand{\arraystretch}{1.2}
\scalebox{0.85}{
\begin{tabular}{@{}l|c|c|c|c|c|c@{}}
\toprule
\textbf{Frames} & \textbf{VR} & \textbf{SC} & \textbf{TO} & \textbf{CE} & \textbf{MU} & \textbf{Avg. Accuracy} \\
\midrule
8  & 43.87\% & 45.30\% & 33.09\% & 37.54\% & 36.78\% & 39.79\% \\
16 & 45.51\% & 50.44\% & 34.03\% & 37.93\% & 40.93\% & 42.12\% \\
32 & 47.66\% & 47.08\% & 34.73\% & 41.35\% & 47.04\% & 43.76\% \\
64 & 50.83\% & 49.25\% & 34.74\% & 41.64\% & 45.11\% & 45.01\% \\
\bottomrule
\end{tabular}
}
\end{table*}

\begin{table*}[htp]
\centering
\caption{{No-Context, Only-Language-Based, and Randomly Language QA Evaluation Trends.} The table presents the performance of the LLama-3.1-70B~\cite{dubey2024llama} model on 4 of 5 models under three evaluation conditions. The results are averaged with predictions across 5 different seeds and a threshold of 0.8.}
\label{tab:meta70b_trends}
\renewcommand{\arraystretch}{1.2}
\scalebox{0.85}{
\begin{tabular}{@{}l|ccccc|c@{}}
\toprule
\textbf{Evaluation Type} & \textbf{VR} & \textbf{SC} & \textbf{TO} & \textbf{CE} & \textbf{MU} & \textbf{Avg Accuracy} \\
\midrule
No-Context (only QA)               & 11.05\% & 11.84\% & 10.85\% & 12.12\% & 10.33\% & 11.22\% \\
+ Language (QA + closed captions) & 11.27\% & 18.53\% & 25.59\% & 14.98\% & 17.68\% & 16.84\% \\
\bottomrule
\end{tabular}
}
\end{table*}

\section{Experiments and Discussion}

We evaluated a range of proprietary and open-source MLLMs on $\mathsf{Cin\acute{e}aste}$ to assess their capabilities in fine-grained movie understanding. Our analysis first presents the main quantitative results, then provides a diagnostic analysis of common model failures, and concludes with ablation studies that validate our benchmark's design.

\subsection{Experimental Setup}
For all open-source models, we provide the maximum number of frames each architecture supports, ensuring a fair comparison of their out-of-the-box capabilities. For proprietary models like GPT-4o, we used 32 frames as a balance between performance, API cost, and maintaining a comparable input resolution to the open-source models being evaluated. Models were tasked with selecting the correct answer from five multiple-choice options. Detailed descriptions of each model's architecture are available in the supplementary material.

\subsection{Main Results: Performance on Cinéaste}
The complete evaluation results are detailed in Table~\ref{tab:vlm_eval}. Our analysis reveals that no model comes close to solving the benchmark; the top-performing model, GPT-4o, reaches 75.89\% accuracy, indicating substantial challenges remain. A clear performance hierarchy exists, with proprietary models from OpenAI (GPT-4o, 75.89\%) and Google (Gemini-2.0-Flash, 66.21\%) leading the evaluation.

The top-tier open-source model, Aria~\cite{li2024aria}, follows at 63.15\%, leveraging an 8x3.5B MoE architecture to significantly surpass standard 7B parameter models, which cluster around 50\% accuracy. This suggests that model architecture and scale are critical factors. However, different models exhibit distinct reasoning strengths. For instance, Aria performs best in Message Understanding (MU) at 67.03\% and Cause and Effect (CE) at 66.21\%, while Gemini-2.0-Flash excels at Visual Reasoning (VR) with 74.28\%.

Temporal Ordering (TO) consistently emerges as a primary bottleneck for most models. The performance gap between GPT-4o (74.26\%) and the next best models in this category, Claude-3.5-Sonnet (63.05\%) and Gemini-2.0-Flash (54.96\%), highlights a significant weakness in long-range sequential reasoning. This significant gap in temporal reasoning underscores that simply processing more frames is insufficient, as both model architecture and its inherent reasoning capabilities are crucial for narrative comprehension.

\subsection{Analysis of Core Failure Modes}
While quantitative scores benchmark performance, a qualitative review of model errors reveals systemic failures in narrative reasoning. We identify three primary patterns that offer a clear roadmap for future research.

First, \textbf{models fail to maintain long-range temporal dependencies.} This is most evident in the \textit{Temporal Ordering} category, where GPT-4o (74.26\%) significantly outperforms capable open-source models like Aria and Gemini-Flash by nearly 20 points (54.78\% and 54.96\%, respectively). This gap suggests a recency bias that fractures narrative coherence; models can sequence adjacent events but fail to connect early-film setups with late-film payoffs. This appears to be a failure of abstract temporal logic, not merely perception, as increasing the visual frame rate provides negligible improvement for this task (see Section 4.4).

Second, \textbf{models fail to resolve conflicting multimodal signals,} often defaulting to literal interpretations of dialogue even when visual cues provide contradictory subtext. This is particularly evident in the nuanced demands of the \textit{Visual Reasoning} category, which was one of GPT-4o's lowest-scoring areas (72.60\%), indicating the difficulty of these tasks even for top models. For example, when asked to identify a non-verbal signal of a character's discomfort, a model might select a plausible but incorrect generic action like "avoiding eye contact" while missing the specific, more subtle visual cue that defines the scene's tension, such as a subtle tremor in a character's hand that betrays their calm dialogue. This indicates a failure in grounding textual information in its visual context.

Finally, \textbf{models struggle to generalize from concrete perception to abstract understanding.} This gap between recognition and comprehension is apparent in the \textit{Message Understanding} category. While top models perform well (GPT-4o: 76.81\%, Aria: 67.03\%), weaker models falter by providing literal descriptions instead of thematic interpretations. For example, when asked what visual motif represents a character's breaking point, a model might identify a visually salient but thematically incorrect object. Its rationale will describe the literal object, demonstrating successful perception but a complete failure at the subsequent cognitive step of symbolic interpretation.

\subsection{Benchmark Validation and Ablation Studies}
To ensure our findings are meaningful, we validated the benchmark's design and robustness.

\textbf{Vision Dependency.} We evaluated a text-only LLaMa-3.1-70B model~\cite{dubey2024llama} under two conditions (Table~\ref{tab:meta70b_trends}). With no context, accuracy was 11.22\% (well below the random chance of 20\%), thereby confirming that our distractor options are effective and questions are not easily guessable. Adding closed captions only improved performance to 16.84\%, proving that visual information is essential. Notably, captions primarily aided Temporal Ordering (+14.7 pts) but provided almost no benefit to Visual Reasoning (+0.2 pts), demonstrating that $\mathsf{Cin\acute{e}aste}$ effectively isolates vision-dependent and text-dependent reasoning skills.

\textbf{Sensitivity to Temporal Context.} An ablation study on VideoLLaMA3~\cite{zhang2025videollama} (Table~\ref{tab:videollama3_ablation}), using only visual frames, shows that average accuracy increases monotonically with more frames (+5.2 pts from 8 to 64 frames). Performance on visually intensive tasks like VR shows the largest gain (+7.0 pts), confirming the benchmark's sensitivity to visual information distributed over time. Full ablation results are available in the supplementary material.

\section{Conclusion}
We presented $\mathsf{Cin\acute{e}aste}$, a benchmark for fine-grained, long-form movie understanding with 3,119 QA pairs across five reasoning categories. Our contribution is twofold: the benchmark itself, created via a robust, automated pipeline with two-stage filtering to ensure multimodal dependency and factual grounding; and a detailed analysis of current MLLMs. Our experiments show that even SOTA models are far from solving these tasks, with top accuracy at 75.89\%. More importantly, our analysis reveals specific, systemic failure modes, including critical deficits in long-range temporal reasoning and multimodal grounding. By diagnosing these weaknesses, $\mathsf{Cin\acute{e}aste}$ provides a clear and challenging roadmap for developing the next generation of models capable of true narrative comprehension.

{
    \small
    \bibliographystyle{ieeenat_fullname}
    \bibliography{main}
}

\clearpage
\setcounter{page}{1}
\setcounter{section}{0}

\section{Dataset Statistics and Distributions}

\begin{figure*}[htp]
    \centering
    \resizebox{0.9\textwidth}{!}{ %
    \begin{minipage}{\textwidth}
        \centering
        \captionsetup[subfigure]{labelformat=empty}  %
        \begin{subfigure}{0.33\textwidth}
            \includegraphics[width=\linewidth]{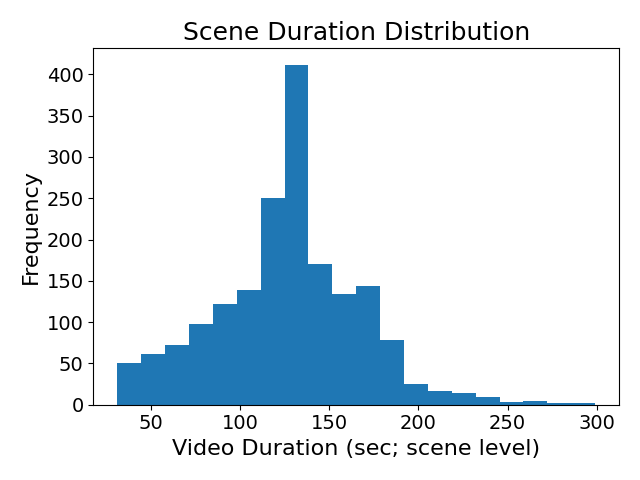}
        \end{subfigure}
        \hspace{-0.5em} %
        \begin{subfigure}{0.33\textwidth}
            \includegraphics[width=\linewidth]{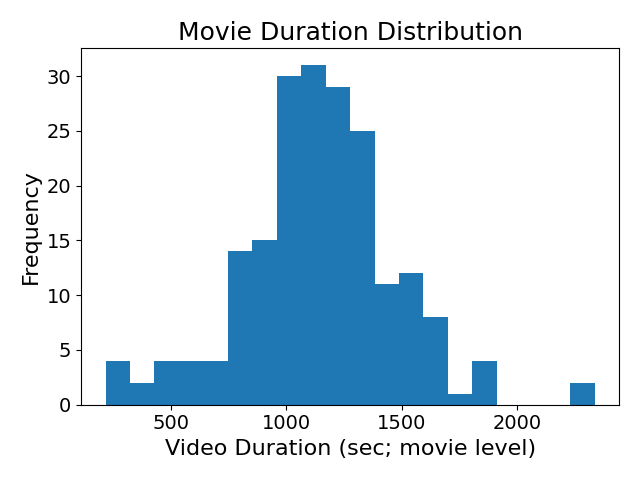}
        \end{subfigure}
        \hspace{-0.5em}
        \begin{subfigure}{0.33\textwidth}
            \includegraphics[width=\linewidth]{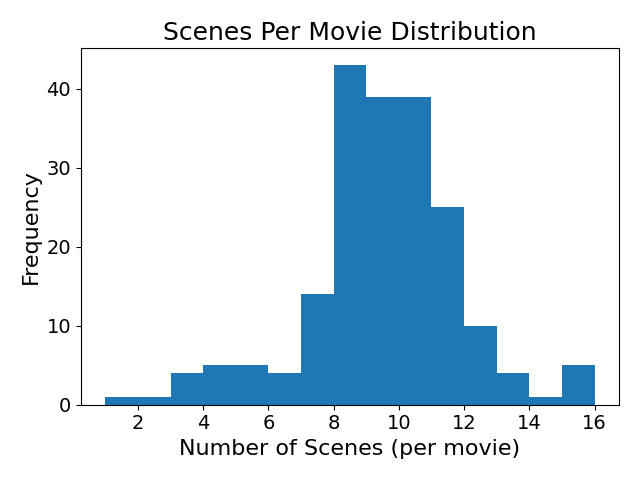}
        \end{subfigure}

        \begin{subfigure}{0.33\textwidth}
            \includegraphics[width=\linewidth]{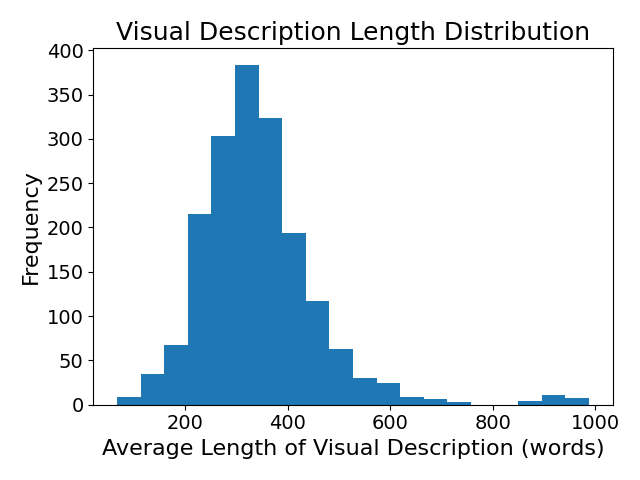}
        \end{subfigure}
        \hspace{-0.5em}
        \begin{subfigure}{0.33\textwidth}
            \includegraphics[width=\linewidth]{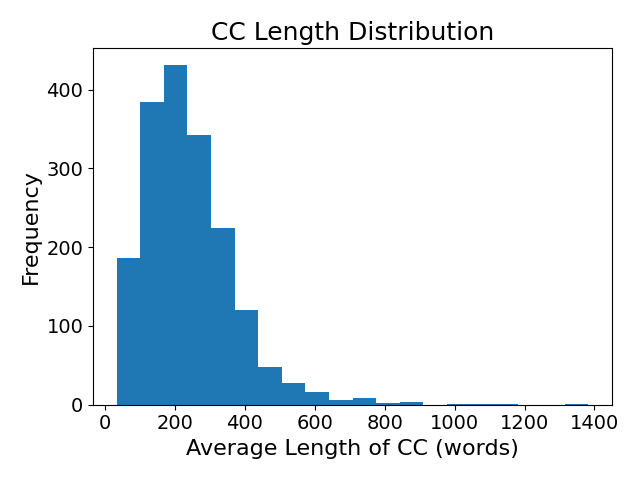}
        \end{subfigure}
        \hspace{-0.5em}
        \begin{subfigure}{0.33\textwidth}
            \includegraphics[width=\linewidth]{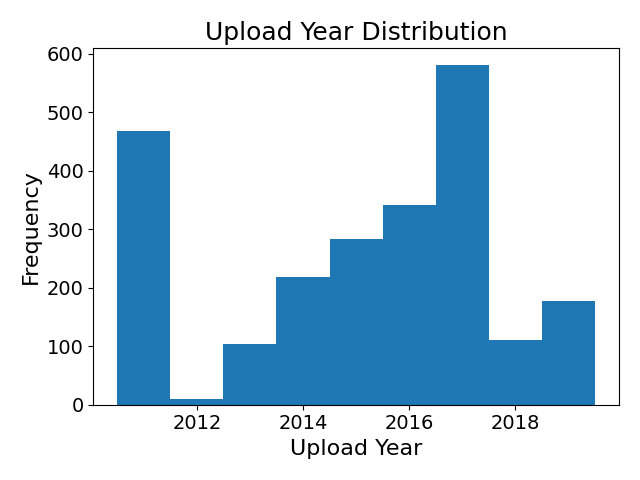}
        \end{subfigure}
    \end{minipage}
    }
\caption{Distribution of QA Pipeline Input Data}
    \label{fig:video_metrics_supp}
    \vspace{-1em}  %
\end{figure*}

This section provides detailed distributions for the input data used in the $\mathsf{Cin\acute{e}aste}$ benchmark generation pipeline. Figure~\ref{fig:video_metrics_supp} illustrates the distributions of scene and movie durations, the number of scenes aggregated per movie, the length of generated visual descriptions and closed captions, and the original upload year of the source video clips. These statistics offer a comprehensive overview of the dataset's scale and characteristics.

\subsection{Model Implementation Details}
We evaluated a range of proprietary and open-source MLLMs to establish a comprehensive performance baseline on $\mathsf{Cin\acute{e}aste}$. All models were evaluated using their publicly available checkpoints and standard inference configurations. For proprietary models, we used their official APIs.

\textbf{Proprietary Models.} We evaluated three leading proprietary models: \textbf{GPT-4o}, \textbf{Gemini-2.0-Flash}, and \textbf{Claude-3.5-Sonnet}. For these models, we followed the video input specifications of their respective APIs. GPT-4o was evaluated using a uniform sampling of 32 frames per video segment.

\textbf{Open-Source Models.} Our evaluation includes a diverse set of open-source models to analyze a range of architectures and training methodologies.
\begin{itemize}
    \item \textbf{Aria}~\cite{li2024aria} is a powerful open-source model featuring a Mixture-of-Experts (MoE) architecture (8x3.5B), which allows for high performance while managing computational load. We evaluated it using its specified input of 128 frames.
    \item \textbf{LongVA-DPO}~\cite{zhang2024longva} and \textbf{LLaVA-NeXT}~\cite{li2024llava} are advanced variants of the LLaVA family. LongVA-DPO is notable for its use of Direct Preference Optimization (DPO) for fine-tuning.
    \item \textbf{VideoLLaMA3}~\cite{zhang2025videollama} is a recent iteration of the VideoLLaMA series, designed for improved video understanding.
    \item Other models, including \textbf{MiniCPM}~\cite{hu2024minicpm}, \textbf{InternVL2}~\cite{chen2024expanding}, and \textbf{ChatUniVi}~\cite{jin2024chat}, represent a variety of other popular 7B parameter MLLM architectures. Unless otherwise specified, these models were evaluated using a uniform sampling of 64 frames.
\end{itemize}

\subsection{Ablation Study on Temporal Sampling Density}
To validate our benchmark's sensitivity to temporal information and to better understand model behavior, we conducted an ablation study on VideoLLaMA3~\cite{zhang2025videollama} by varying the number of input frames. The full results are presented in Table~\ref{tab:supp_videollama3_ablation}. The evaluation was performed without providing closed captions to isolate the impact of visual context.

The results show that average accuracy improves monotonically as the frame count increases from 8 to 64, confirming that denser visual information is generally beneficial. Performance on visually intensive tasks like \textit{Visual Reasoning} (VR) shows the most significant and consistent improvement (+7.0 pts), as more frames provide more opportunities for object and scene recognition. In contrast, the gain in \textit{Temporal Ordering} (TO) is minimal (+1.7 pts), reinforcing our main paper's finding that the core challenge in this category is one of abstract logical reasoning, not merely a lack of visual evidence. Interestingly, performance in categories like \textit{State Changes} (SC) and \textit{Message Understanding} (MU) appears non-monotonic, peaking at 16 and 32 frames, respectively. This may suggest that for certain abstract tasks, an excess of visual frames can introduce noise or distract the model, indicating a task-specific optimal range for temporal sampling density.

\begin{table*}
\centering
\caption{
    \textbf{Ablation Study of VideoLLaMA3 with Increasing Frame Counts.} The table shows accuracy on $\mathsf{Cin\acute{e}aste}$ using only video frames.
}
\label{tab:supp_videollama3_ablation}
\renewcommand{\arraystretch}{1.2}
\scalebox{0.9}{
\begin{tabular}{@{}l|ccccc|c@{}}
\toprule
\textbf{Frames} & \textbf{VR} & \textbf{SC} & \textbf{TO} & \textbf{CE} & \textbf{MU} & \textbf{Avg. Accuracy} \\
\midrule
8  & 43.87\% & 45.30\% & 33.09\% & 37.54\% & 36.78\% & 39.79\% \\
16 & 45.51\% & \textbf{50.44\%} & 34.03\% & 37.93\% & 40.93\% & 42.12\% \\
32 & 47.66\% & 47.08\% & 34.73\% & 41.35\% & \textbf{47.04\%} & 43.76\% \\
64 & \textbf{50.83\%} & 49.25\% & \textbf{34.74\%} & \textbf{41.64\%} & 45.11\% & \textbf{45.01\%} \\
\bottomrule
\end{tabular}
}
\end{table*}

\end{document}